\documentclass[11pt,a4paper]{article}
\usepackage[hyperref]{emnlp2020}
\usepackage{times}
\usepackage{latexsym}
\usepackage{booktabs}
\usepackage{mwe}

\usepackage{subcaption}

\usepackage{listings}
\usepackage{xcolor}
\usepackage{lipsum}
\definecolor{codegreen}{rgb}{0,0.6,0}
\definecolor{codegray}{rgb}{0.5,0.5,0.5}
\definecolor{codepurple}{rgb}{0.58,0,0.82}
\definecolor{backcolour}{rgb}{0.95,0.95,0.92}

\lstdefinestyle{mystyle}{
    backgroundcolor=\color{backcolour},   
    commentstyle=\color{codegreen},
    keywordstyle=\color{magenta},
    numberstyle=\tiny\color{codegray},
    stringstyle=\color{codepurple},
    basicstyle=\ttfamily\footnotesize,
    breakatwhitespace=false,         
    breaklines=false,                 
    captionpos=b,                    
    keepspaces=true,                 
    numbers=left,                    
    numbersep=5pt,                  
    showspaces=false,                
    showstringspaces=false,
    showtabs=false,                  
    tabsize=2
}
\usepackage{microtype}

\aclfinalcopy % Uncomment this line for the final submission
\DeclareRobustCommand{\huggingface}{%
  \begingroup\normalfont
  \vspace{-0.2em}%
  \raisebox{-0.4em}{%
  \includegraphics[height=1.5em]{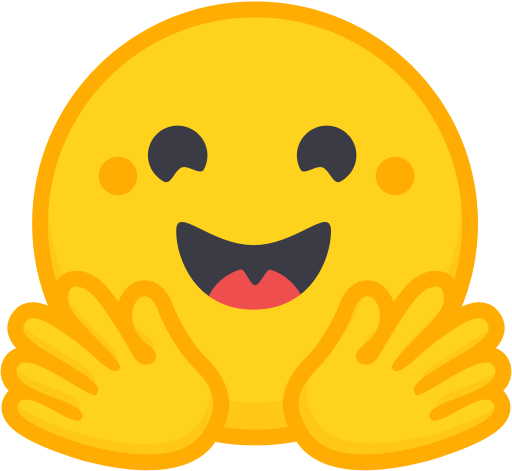}%
  }%
  \kern 0.4em%
  \endgroup
}
\title{\huggingface Transformers: State-of-the-Art Natural Language Processing}

 \author{ \vspace{-0.5cm} \\
 Thomas~Wolf,
 Lysandre~Debut,
 Victor~Sanh,
 Julien~Chaumond,   
 Clement~Delangue, \\
 Anthony~Moi,
 Pierric~Cistac,
 Tim~Rault,  
 R\'{e}mi~Louf,
 Morgan~Funtowicz, Joe~Davison, \\
 Sam~Shleifer, Patrick~von~Platen, Clara~Ma, Yacine~Jernite, Julien~Plu, Canwen~Xu, \\
 Teven~Le~Scao, Sylvain~Gugger, Mariama~Drame, Quentin~Lhoest, Alexander~M.~Rush \\
 \\
 Hugging Face, Brooklyn, USA /  \texttt{\{first-name\}@huggingface.co}}

\date{}

\begin{document}
\maketitle
\begin{abstract}

    Recent progress in natural language processing has been driven by 
    advances in both model architecture and model pretraining. 
    Transformer architectures have facilitated building higher-capacity models and pretraining has made it possible to effectively utilize this capacity for a wide variety of tasks. \textit{Transformers} is an open-source library with the goal of 
    opening up these advances to the wider machine learning community. The library consists of carefully engineered state-of-the art Transformer architectures under a unified API. Backing this library is a curated collection of pretrained models made by and available for the community. \textit{Transformers} is designed to be extensible by researchers, simple for practitioners, and fast and robust in industrial deployments.  The library is available at \url{https://github.com/huggingface/transformers}.

%    Still, creating these general-purpose models remains an expensive and time-consuming process restricting the use of these methods to a small sub-set of the wider NLP community. In this paper, we present HuggingFace's Transformers library, a library for state-of-the-art NLP, making these developments available to the community by gathering state-of-the-art general-purpose pretrained models under a unified API together with an ecosystem of libraries, examples, tutorials and scripts targeting many downstream NLP tasks. Transformers library features carefully crafted model implementations and high-performance pretrained weights for two main deep learning frameworks, PyTorch and TensorFlow, while supporting all the necessary tools to analyze, evaluate and use these models in downstream tasks such as text/token classification, questions answering and language generation among others. The library has gained significant organic traction and adoption among both the researcher and practitioner communities. We are committed at HuggingFace to pursue the efforts to develop this toolkit with the ambition of creating the standard library for building NLP systems. Transformers library is available at \url{https://github.com/huggingface/transformers}.

\end{abstract}

% What problem does the proposed system address?
% Why is the system important and what is its impact?
% What is the novel in the approach/technology on which this system is based?
% Who is the target audience?
% How does the system work?
% How does it compare with existing systems?
% How is the system licensed?

\section{Introduction}

The Transformer \cite{Vaswani2017-bg} has rapidly become the dominant architecture for natural language processing, surpassing alternative neural models such as convolutional  and recurrent neural networks in performance for tasks in both natural language understanding and natural language generation. The architecture scales with training data and model size, facilitates efficient parallel training, and captures long-range sequence features.

Model pretraining~\cite{McCann2017-uz,Howard2018UniversalLM,Peters2018-jy,Devlin2018-gk} allows models to be trained on generic corpora and subsequently be easily adapted to specific tasks with strong performance. The Transformer architecture is particularly conducive to pretraining on large text corpora, leading to major gains in accuracy on downstream tasks including text classification \citep{Yang2019XLNetGA}, language understanding \citep{Liu2019RoBERTaAR, Wang2018GLUEAM, Wang2019SuperGLUEAS}, machine translation \citep{Lample2019CrosslingualLM}, coreference resolution \citep{joshi2019spanbert}, commonsense inference \citep{Bosselut2019COMETCT}, and summarization~\cite{Lewis2019-ak} among others.

This advance leads to a wide range of practical challenges that must be addressed in order for these models to be widely utilized. 
The ubiquitous use of the Transformer calls for systems to train, analyze, scale, and augment the model on a variety of platforms. The architecture is used as a building block to design increasingly sophisticated extensions and precise experiments. The pervasive adoption of pretraining methods has led to the need to distribute, fine-tune, deploy, and compress the core pretrained models used by the community.
%Practitioners increasingly rely on these as standardized starting points and benchmarks for further comparison. 

\begin{figure*}
\centering
\hspace*{-1cm}\includegraphics[width=1.1\textwidth]{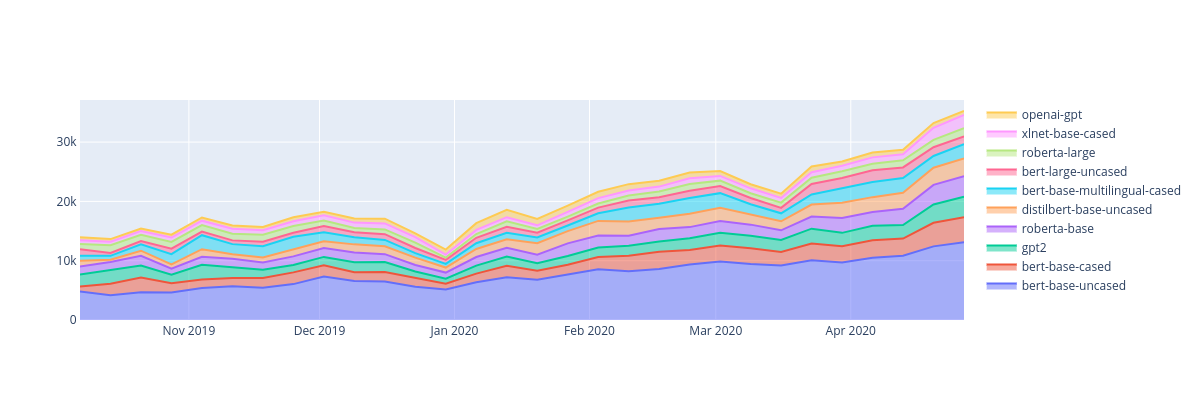}
\caption{Average daily unique downloads of the most downloaded pretrained models, Oct. 2019 to May 2020.  }
\label{fig:downloads}
\end{figure*} 

\textit{Transformers} is a library dedicated to supporting Transformer-based architectures and facilitating the distribution of pretrained models.
At the core of the libary is an implementation of the Transformer which is designed for both research and production. The philosophy is to support industrial-strength implementations of popular model variants that are easy to read, extend, and deploy. On this foundation, the library supports the distribution and usage of a wide-variety of pretrained models in a centralized model hub. This hub supports users to compare different models with the same minimal API and to experiment with shared models on a variety of different tasks.

\textit{Transformers} is an ongoing effort maintained by the team of engineers and researchers at Hugging Face with support from a vibrant community of over 400 external contributors. The library is released under the Apache 2.0 license and is available on GitHub\footnote{\url{https://github.com/huggingface/transformers}}. Detailed documentation and tutorials are available on Hugging Face's website\footnote{\url{https://huggingface.co/transformers/}}.

%In this technical report, we discuss each aspect of Transformers and how it relates to other tools for general-purpose NLP. We begin by discussing related libraries. We then discuss the core internals of the library and its design, the usage and goals of the community model hub, and how the library facilitates deployment in practice.   

%We are committed to the twin efforts of developing the library and fostering positive interaction among its community members, with the ambition of creating the standard library for modern deep learning NLP.

\section{Related Work}

The NLP and ML communities have a strong culture of building open-source research tools.  
%While Transformers is designed for a variety of NLP tasks, its main neural structure is most related to systems for language modeling and translation, under which tools like Transformer and BERT were developed. 
The structure of \textit{Transformers} is inspired by the pioneering tensor2tensor library~\citep{tensor2tensor} and the original source code for BERT~\citep{Devlin2018-gk}, both from Google Research. The concept of providing easy caching for pretrained models stemmed from AllenNLP~\cite{Gardner2018-tg}. 
The library is also closely related to neural translation and language modeling systems, such as Fairseq~\cite{Ott2019-pi},  OpenNMT~\cite{klein-etal-2017-opennmt}, Texar~\cite{Hu2018-xj}, Megatron-LM~\cite{shoeybi2019megatron}, and Marian NMT~\cite{Junczys-Dowmunt2018-xq}. Building on these elements, \textit{Transformers} adds extra user-facing features to allow for easy downloading, caching, and fine-tuning of the models as well as seamless transition to production. \textit{Transformers} maintains some compatibility with these libraries, most directly including a tool for performing inference using models from Marian NMT and Google's BERT.

There is a long history of easy-to-use, user-facing libraries for general-purpose NLP. Two core libraries are NLTK~\cite{Loper2002-xd} and Stanford CoreNLP~\cite{Manning2014-pc}, which collect a variety of different approaches to NLP in a single package. 
More recently, general-purpose, open-source libraries have focused primarily on machine learning for a variety of NLP tasks, these include Spacy~\cite{Honnibal2017-bx}, AllenNLP~\cite{Gardner2018-tg}, flair~\cite{Akbik2019-ch}, and Stanza~\cite{Qi2020-he}. \textit{Transformers} provides similar functionality as these libraries. Additionally, each of these libraries now uses the \textit{Transformers} library and model hub as a low-level framework.

Since \textit{Transformers} provides a hub for NLP models, it is also related to popular model hubs including Torch Hub~\nocite{noauthor_undated-qt} and TensorFlow Hub~\nocite{noauthor_undated-sy} which collect framework-specific model parameters for easy use. Unlike these hubs, \textit{Transformers} is domain-specific which allows the system to provide automatic support for model analysis, usage, deployment, benchmarking, and easy replicability.

\section{Library Design}

\begin{figure*}
\begin{center}
    \begin{tabular}{lccp{3cm}p{3cm}}
    \toprule
    \multicolumn{5}{c}{\textbf{Heads}}\\
          Name & Input & Output & Tasks & Ex. Datasets  \\
    \midrule
         Language Modeling & $x_{1:n-1}$ & $x_n \in {\cal V}$ & Generation & WikiText-103\\
         Sequence Classification & $x_{1:N}$ & $y \in {\cal C}$ & Classification, \newline Sentiment Analysis & GLUE, SST, \newline MNLI \\ 
         Question Answering & $x_{1:M},$ \newline $x_{M:N}$ & $y$ span $[1:N]$ & QA,  Reading\newline Comprehension & SQuAD, \newline Natural Questions \\
         Token Classification & $x_{1:N}$ & $y_{1:N} \in {\cal C}^N$ &   NER, Tagging &  OntoNotes, WNUT \\
         Multiple Choice & $x_{1:N}, {\cal X}$ & $y \in {\cal X}$ & Text Selection & SWAG, ARC \\
         Masked LM & $x_{1:N\setminus n}$ & $x_n \in {\cal V}$ & Pretraining & Wikitext, C4 \\
         Conditional Generation & $x_{1:N}$ & $y_{1:M} \in {\cal V}^M$ & Translation,\newline Summarization & WMT, IWSLT, \newline CNN/DM, XSum \\
    \bottomrule
    \end{tabular}
    \end{center}
    \vspace*{0.8cm}
    
    \begin{minipage}[t]{\columnwidth}
    \begin{tabular}{lr}
    \toprule
         \multicolumn{2}{c}{\textbf{Transformers}}  \\
    \midrule
    \multicolumn{2}{c}{Masked $[x_{1:N \setminus n} \Rightarrow x_n]$}\\
    \midrule
     BERT &  \cite{Devlin2018-gk}\\ 
     RoBERTa &  \cite{Liu2019-rl} \\  
    \midrule
    \multicolumn{2}{c}{Autoregressive  $[x_{1:n-1} \Rightarrow x_n]$ }\\
    \midrule
     GPT / GPT-2 & \cite{Radford2019-lx} \\
     Trans-XL & \cite{Dai2019-ig} \\
     XLNet & \cite{Yang2019XLNetGA} \\ 
    \midrule
    \multicolumn{2}{c}{Seq-to-Seq $[\sim x_{1:N} \Rightarrow x_{1:N}]$ }\\
    \midrule
     BART & \cite{Lewis2019-ak} \\
     T5 & \cite{Raffel2019-sk} \\ 
     MarianMT & (\textcolor{darkblue}{J.-Dowmunt et al., 2018})\nocite{Junczys-Dowmunt2018-xq} \\ 
    \midrule
     \multicolumn{2}{c}{Specialty: Multimodal} \\ 
    \midrule
     MMBT & \cite{Kiela2019-zl} \\ 
    \midrule
    \multicolumn{2}{c}{Specialty: Long-Distance}  \\
    \midrule
     Reformer & \cite{Kitaev2020-cn} \\ 
     Longformer & \cite{Beltagy2020-me} \\
    \midrule
    \multicolumn{2}{c}{Specialty: Efficient} \\ 
    \midrule
    ALBERT & \cite{Lan2019-oe}\\
    Electra & \cite{Clark2020-rd} \\
    DistilBERT & \cite{Sanh2019-ci}\\
    \midrule
    \multicolumn{2}{c}{Specialty: Multilingual} \\ 
    \midrule
    XLM/RoBERTa & \cite{Lample2019-hr}\\

    \bottomrule
    \end{tabular}
    
    \end{minipage}
    \hfill
    \begin{minipage}[t]{\columnwidth}
    \centering 
        
        \includegraphics[width=0.85\linewidth]{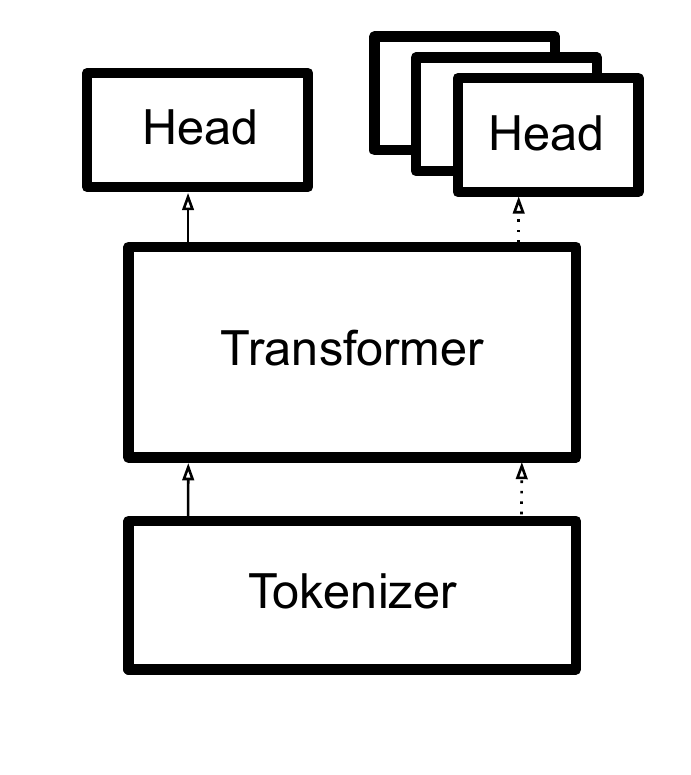}
    
    \vspace{1cm}
    \begin{tabular}{ll}
    
    \toprule
        \multicolumn{2}{c}{\textbf{Tokenizers}} \\
    \midrule
    Name & Ex. Uses \\ 
    \midrule
    Character-Level BPE & NMT, GPT \\ 
    Byte-Level BPE & GPT-2\\ 
    WordPiece &  BERT \\
    SentencePiece &  XLNet \\
    Unigram & LM \\
    Character & Reformer \\
    Custom & Bio-Chem \\
    \bottomrule
    \end{tabular}
    \end{minipage}

    \caption{The \textit{Transformers} library. \textbf{(Diagram-Right)} Each model is made up of a Tokenizer, Transformer, and Head. The model is pretrained with a fixed head and can then be further fine-tuned with alternate heads for different tasks. \textbf{(Bottom)} Each model uses a specific Tokenizer either implemented in Python or in Rust. These often differ in small details, but need to be in sync with pretraining. \textbf{(Left)}  Transformer architectures specialized for different tasks, e.g. understanding versus generation, or for specific use-cases, e.g. speed, image+text. 
    \textbf{(Top)} heads allow a Transformer to be used for different tasks. Here we assume the input token sequence is $x_{1:N}$ from a vocabulary $\cal V$, and $y$ represents different possible outputs, possibly from a class set $\cal C$. Example datasets represent a small subset of example code distributed with the library. }
    \label{fig:transformers}
\end{figure*}

\textit{Transformers} is designed to mirror the standard NLP machine learning model pipeline: process data, apply a model, and make predictions. Although the library includes tools facilitating training and development, in this technical report we focus on the core modeling specifications. For complete details about the features of the library refer to the documentation available on \url{https://huggingface.co/transformers/}.

Every model in the library is fully defined by three building blocks shown in the diagram in Figure~\ref{fig:transformers}: (a) a tokenizer, which converts raw text to sparse index encodings, (b) a transformer, which transforms sparse indices to contextual embeddings, and (c) a head, which uses contextual embeddings to make a task-specific prediction. Most user needs can be addressed with these three components.

\paragraph{Transformers} Central to the library are carefully tested implementations of Transformer architecture variants which are widely used in NLP. The full list of 
currently implemented architectures is shown in Figure~\ref{fig:transformers}~(Left). While each of these architectures shares the same multi-headed attention core, there are significant differences between them including positional representations, masking, padding, and the use of sequence-to-sequence design. Additionally, various models are built to target different applications of NLP such as understanding, generation, and conditional generation, plus specialized use cases such as fast inference or multi-lingual applications. 

Practically, all models follow the same hierarchy of abstraction: a base class implements the model's computation graph from an encoding (projection on the embedding matrix) through the series of self-attention layers to the final encoder hidden states. The base class is specific to each model and closely follows the model's original implementation which gives users the flexibility to easily dissect the inner workings of each individual architecture. In most cases, each model is implemented in a single file to enable ease of extensibility.

Wherever possible, different architectures follow the same API allowing users to switch easily between different models. A set of \texttt{Auto} classes provides a unified API that enables very fast switching between models and even between frameworks. These classes automatically instantiate with the configuration specified by the user-specified pretrained model.

\paragraph{Tokenizers} A critical NLP-specific aspect of the library is the implementations of the tokenizers necessary to use each model. Tokenizer classes (each inheriting from a common base class) can either be instantiated from a corresponding pretrained model or can be configured manually. These classes store the vocabulary token-to-index map for their corresponding model and handle the encoding and decoding of input sequences according to a model's specific tokenization process. The tokenizers implemented are shown in Figure~\ref{fig:transformers}~(Right).  Users can easily modify tokenizer with interfaces to add additional token mappings, special tokens (such as classification or separation tokens), or otherwise resize the vocabulary.

Tokenizers can also implement additional useful features for the users. These range from token type indices in the case of sequence classification to maximum length sequence truncating taking into account the added model-specific special tokens (most pretrained Transformer models have a maximum sequence length).

For training on very large datasets, Python-based tokenization is often undesirably slow. In the most recent release, \textit{Transformers}  switched its implementation to use a highly-optimized tokenization library by default. This low-level library, available at \url{https://github.com/huggingface/tokenizers}, is written in Rust to speed up the tokenization procedure both during training and deployment.

%For training on very large datasets, Python-based tokenization can be the bottleneck on training speed and so this is an important practical extension.

\paragraph{Heads} Each Transformer can be paired with one out of several ready-implemented 
heads with outputs amenable to common types of tasks. These heads are implemented as additional wrapper classes on top of the base class, adding a specific output layer, and optional loss function, on top of the Transformer's contextual embeddings. The full set of implemented heads are shown in Figure~\ref{fig:transformers}~(Top). These classes follow a similar naming pattern: \texttt{XXXForSequenceClassification} where \texttt{XXX} is the name of the model and can be used for adaptation (fine-tuning) or pretraining. Some heads, such as conditional generation,  support extra functionality like sampling and beam search. 

For pretrained models, we release the heads used to pretrain the model itself. For instance, for BERT we release the language modeling and  next sentence prediction heads which allows easy for adaptation using the pretraining objectives. We also make it easy for users to utilize the same core Transformer parameters with a variety of other heads for finetuning. While each head can be used generally, the library also includes a collection of examples that show each head on real problems. These examples demonstrate how a pretrained model can be adapted with a given head to achieve state-of-the-art results on a large variety of NLP tasks.

\section{Community Model Hub}

\begin{figure*}
\begin{subfigure}[b]{0.5\textwidth}
    \centering
        \includegraphics[width=0.9\linewidth]{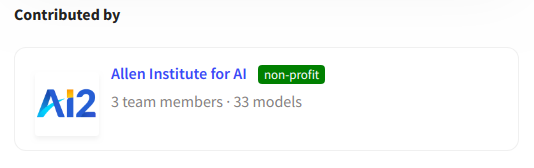}\\
        \includegraphics[width=0.9\linewidth]{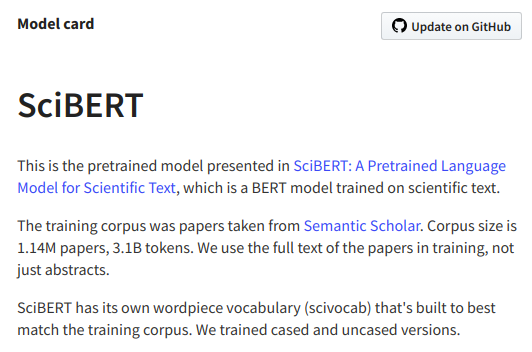}
\end{subfigure}  
\begin{subfigure}[b]{0.5\textwidth}
    \centering
    \includegraphics[width=0.9\linewidth]{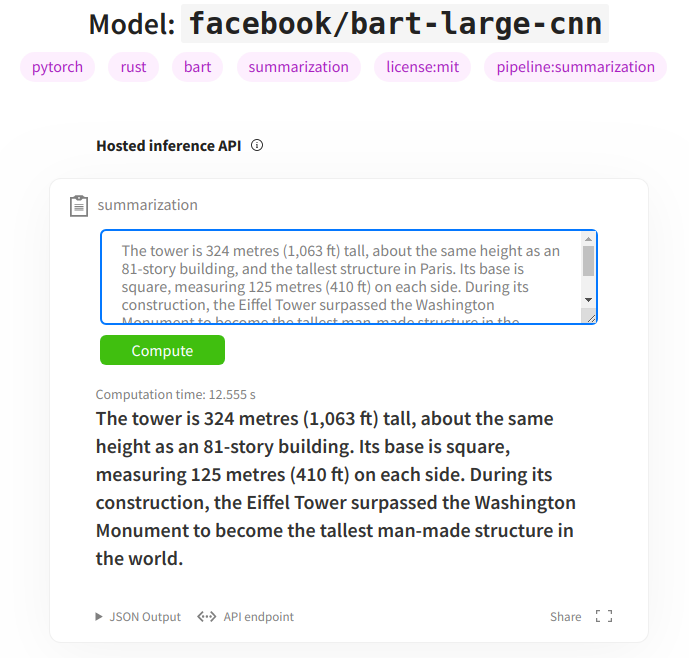}
\end{subfigure}
\caption{\textit{Transformers} Model Hub. \textbf{(Left)} Example of a model page and model card for SciBERT~\cite{Beltagy2019-op}, a pretrained model targeting extraction from scientific literature submitted by a community contributor. \textbf{(Right)} Example of an automatic inference widget for the pretrained BART~\cite{Lewis2019-ak} model for summarization. Users can enter arbitrary text and a full version of the model is deployed on the fly to produce a summary. }
\label{fig:hub}
\end{figure*}

%While model architecture is important for Transformers usage, the main driver of model accuracy in recent years has been the use of pretraining on large-amounts of unsupervised data \cite{}. 
\textit{Transformers} aims to facilitate easy use and distribution of pretrained models. 
Inherently this is a community process; a single pretraining run facilitates fine-tuning on many specific tasks. The Model Hub makes it simple for any end-user to access a model for use with their own data. This hub now contains 2,097 user models, both pretrained and fine-tuned, from across the community. Figure~\ref{fig:downloads} shows the increase and distribution of popular transformers over time. While core models like BERT and GPT-2 continue to be popular, other specialized models including  DistilBERT~\cite{Sanh2019-ci}, which was developed for the library, are now widely downloaded by the community.

%A goal of the Transformers as a library is to facilitate this process on both ends. For architects, it provides a method of model distribution and engagement. For trainers, it provides a source of raw pretrained models and a method for distributing fine-tuned models. For users, it provides a way to try out new approaches either directly on the web or through simple notebooks. 

The user interface of the Model Hub is designed to be simple and open to the community. To upload a model, 
any user can sign up for an account and use a command-line interface to produce an archive consisting a tokenizer, transformer, and head. This bundle may be a model trained through the library or converted from a checkpoint of other popular training tools. These models are then stored and given a canonical name which a user can use to download, cache, and run the model either for fine-tuning or inference in two lines of code. To load FlauBERT~\cite{le2020flaubert}, a BERT model pretrained on a French training corpus, the command is:

\begin{lstlisting}[language=python]
tknzr = AutoTokenizer.from_pretrained(
    "flaubert/flaubert_base_uncased")
model = AutoModel.from_pretrained(
    "flaubert/flaubert_base_uncased")
\end{lstlisting}

When a model is uploaded to the Model Hub, it is automatically given a landing page describing its core properties, architecture, and use cases. Additional model-specific metadata can be provided via a model card~\cite{Mitchell2018-ai} that describes properties of its training, a citation to the work, datasets used during pretraining, and any caveats about known biases in the model and its predictions. An example model card is shown in Figure~\ref{fig:hub}~(Left).

Since the Model Hub is specific to transformer-based models, we can target use cases that would be difficult for more general model collections.
For example, because each uploaded model includes metadata concerning its structure, 
the model page can include live inference that allows users to experiment with output of models on a real data. Figure~\ref{fig:hub}~(Right) shows an example of the model page with live inference. Additionally, model pages include links to other model-specific tools like benchmarking and visualizations. For example, model pages can link to exBERT~\cite{Hoover2019-kc}, a Transformer visualization library. 

%Researchers have used it as a way to study the structure of transformer attention heads~\cite{Michel2019-fh} and benchmark different models~\cite{}.

\paragraph{Community Case Studies} The Model Hub highlights how \textit{Transformers} is used by a variety of different community stakeholders. We summarize three specific observed use-cases in practice. We highlight specific systems developed by users with different goals following the  architect, trainer, and end-user distinction of~\citet{strobelt2017lstmvis}: 

%The ecosystem of deep learning based NLP includes a variety of different participants, spanning from those utilizing massive amounts of compute to train new models to those fine-tuning models on new datasets to those simply deploying fixed models in applications. At a high-level, we can classify these roles.

%a) \textit{architects}, those developing and training new architectures, b) \textit{trainers}, those utilizing pretrained models to fit to new data, c) \textit{users}, those deploying and utilizing fully trained models. We begin by discussing three case studies of how these practitioners have used Transformers. 

\noindent \textit{Case 1: Model Architects} AllenAI, a major NLP research lab, developed a new pretrained model for improved extraction from biomedical texts called SciBERT~\cite{Beltagy2019-op}. They were able to train the model utilizing data from PubMed to produce a masked language model with state-of-the-art results on targeted text. They then used the Model Hub to distribute the model and promote it as part of their CORD - COVID-19 challenge, making it trivial for the community to use.

\noindent \textit{Case 2: Task Trainers} Researchers at NYU were interested in developing a test bed for the performance of \textit{Transformers} on a variety of different semantic recognition tasks. Their framework Jiant~\cite{pruksachatkun2020jiant} allows them to experiment with different ways of pretraining models and comparing their outputs. They used the \textit{Transformers} API as a generic front-end and performed fine-tuning on a variety of different models, leading to research on the structure of BERT~\cite{Tenney2019BERTRT}.

\noindent \textit{Case 3: Application Users} Plot.ly, a company focused on user dashboards and analytics, was interested in deploying a model for automatic document summarization. They wanted an approach that scaled well and was simple to deploy, but had no need to train or fine-tune the model. They were able to search the Model Hub and find \textit{DistilBART}, a pretrained and fine-tuned summarization model designed for accurate, fast inference. They were able to run and deploy the model directly from the hub with no required research or ML expertise. 

\section{Deployment}

An increasingly important goal of \textit{Transformers} is to make it easy to efficiently deploy model to production. Different users have different production needs, and deployment often requires solving significantly different challenges than training. The library thereforce allows for several different strategies for production deployment.

One core propery of the libary is that models are available both in PyTorch and TensorFlow, and there is interoperability between both frameworks. A model trained in one of frameworks can be saved through standard serialization and be reloaded from the saved files in the other framework seamlessly. This makes it particularly easy to switch from one framework to the other one along the model life-time (training, serving, etc.).

Each framework has deployment recommendations. For example, in PyTorch, models are compatible with TorchScript, an intermediate representation of a PyTorch model that can then be run either in Python in a more efficient way, or in a high-performance environment such as C++. Fine-tuned models can thus be exported to production-friendly environment, and run through TorchServing. TensorFlow includes several serving options within its ecosystem, and these can be used directly. 

\begin{figure}
    \centering
    \includegraphics[width=\linewidth]{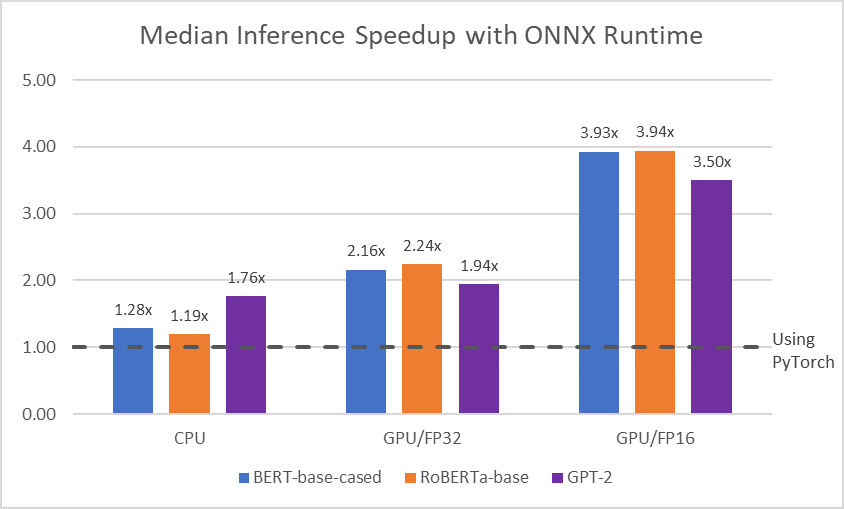}
    \caption{Experiments with \textit{Transformers} inference in collaboration with ONNX.}
    \label{fig:onnx}
\end{figure}

\textit{Transformers} can also export models to intermediate neural network formats for further compilation. It supports converting models to the Open Neural Network Exchange format (ONNX) for deployment. Not only does this allow the model to be run in a standardized interoperable format, but also leads to significant speed-ups. Figure~\ref{fig:onnx} shows experiments run in collaboration with the ONNX team to optimize BERT, RoBERTa, and GPT-2 from the \textit{Transformers} library. Using this intermediate format, ONNX was able to achieve nearly a 4x speedup on this model. The team is also experimenting with other promising intermediate formats such as JAX/XLA~\cite{jax2018github} and  TVM~\cite{chen2018tvm}.

Finally, as Transformers become more widely used in all NLP applications, it is increasingly important to deploy to edge devices such as phones or home electronics. Models can use adapters to convert models to \textit{CoreML} weights that are suitable to be embedded inside a iOS  application, to enable on-the-edge machine learning. Code is also made available\footnote{\url{https://github.com/huggingface/swift-coreml-transformers}}. Similar methods can be used for Android devices.

%While Transformers is designed to facilitate simple training and fine-tuning, as these models leave the research lab, it is increasingly important for them to be utilized efficiently and robustly in production deployment environments. The library includes extensions to make it easy to utilize the models in these contexts. 

%Optimizing large machine learning models for production is an ongoing effort in the community and there are many current engineering efforts towards that goal. The distillation of large models (e.g. \textit{DistilBERT} \citep{sanh2019distilbert}) is one of the most promising directions. It lets users of \texttt{Transformers} run more efficient versions of the models, even with strong latency constraints and on inexpensive CPU servers. 

\section{Conclusion}

As Transformer and pretraining play larger roles in NLP, it is important for these models to be accessible to researchers and end-users. \textit{Transformers} is an open-source library and community designed to facilitate users to access large-scale pretrained models, to build and experiment on top of them, and to deploy them in downstream tasks with state-of-the-art performance. \textit{Transformers} has gained significant organic traction since its release and is set up to continue to provide core infrastructure while helping to facilitate access to new models.

%We are committed to supporting this community and 

\bibliographystyle{acl_natbib}
\bibliography{transformers}%,anthology}

\end{document}